\documentclass[letterpaper, 10 pt, conference]{ieeeconf}  
\usepackage{amsmath} 
\usepackage{graphicx}
\usepackage{amsmath}
\usepackage{amssymb}
\usepackage{mathrsfs} %
\usepackage{xcolor}
\usepackage{url}
\usepackage{caption}
\usepackage{booktabs} 
\usepackage{tabularray} 
\usepackage{tabularx} 
\usepackage{ragged2e} 
\usepackage{booktabs} 
\usepackage{cuted}
\usepackage{array}      %
\usepackage{multirow}   %
\usepackage{booktabs}   %
\usepackage[utf8]{inputenc}
\usepackage{svg} 
\usepackage{amssymb}
\usepackage{algpseudocode}  
\usepackage{amsmath}  
\usepackage{booktabs}
\usepackage[colorlinks,linkcolor=blue]{hyperref}

\IEEEoverridecommandlockouts                              

\title{\LARGE \bf 
\vspace{-20pt}
Quadruped robot traversing 3D complex environments 

with limited perception
}

\author{Yi Cheng*$^{1}$, Hang Liu*$^{2}$, Guoping Pan$^{1}$, Linqi Ye†$^{3}$, Houde Liu†$^{1}$, Bin Liang$^{1}$\\\href{quad-traverse-go2.github.io}{Quad-Traverse-Go2.github.io}
\thanks{* Equal Contributions}
\thanks{† corresponding author}
\thanks{Research supported by the National Natural Science Foundation of China under grants No.92248304 and Shenzhen Science Fund for Distinguished Young Scholars under Grant RCJC20210706091946001}
\thanks{$^{1}$ Tsinghua University, 100084 Beijing, China }%
\thanks{$^{2} $ University of Michigan, Ann Arbor, MI 48109, USA}
\thanks{$^{3} $ Shanghai University, 200444 Shanghai, China. }
}

\begin{document}

\let\oldtwocolumn\twocolumn

\renewcommand\twocolumn[1][]{%
    \oldtwocolumn[{#1}{
    \begin{flushleft}
        \centering
        \vspace{-20pt}
           
        \includegraphics[clip,trim=0cm 0cm 0cm 0cm,width=0.98\textwidth]{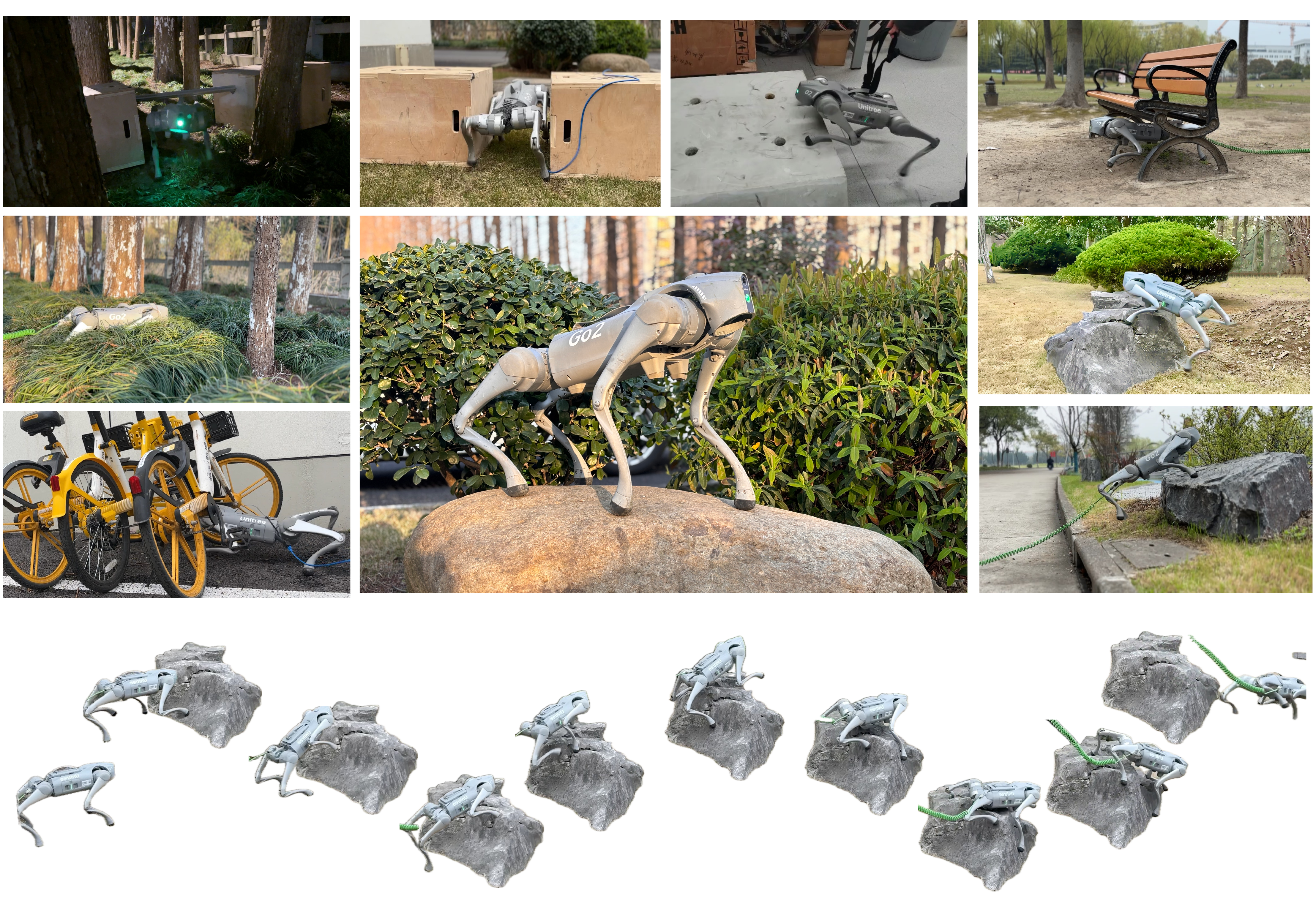}
        
        \captionsetup{justification=justified} 
        
        \captionof{figure}{
             We tested the passability of the quadruped robot in more than 20 kinds of 3D complex environments, indoor and outdoor, without any external sensing devices such as radar, camera, etc., and our method showed good performance. In the lower part of the image is an example of a quadruped robot traversing highland. It is worth noting that the obstacles are unstructured and the robot can still traverse normally.
        }\label{fig:setup}
    \end{flushleft}
    }]
}

\maketitle
\thispagestyle{empty}
\pagestyle{empty}

\begin{abstract}
Traversing 3-D complex environments has always been a significant challenge for legged locomotion.  Existing methods typically rely on external sensors such as vision and lidar to preemptively react to obstacles by acquiring environmental information.  However, in scenarios like nighttime or dense forests, external sensors often fail to function properly, necessitating robots to rely on proprioceptive sensors to perceive diverse obstacles in the environment and respond promptly.  This task is undeniably challenging.  Our research finds that methods based on collision detection can enhance a robot's perception of environmental obstacles.  In this work, we propose an end-to-end learning-based quadruped robot motion controller that relies solely on proprioceptive sensing.  This controller can accurately detect, localize, and agilely respond to collisions in unknown and complex 3D environments, thereby improving the robot's traversability in complex environments.  We demonstrate in both simulation and real-world experiments that our method enables quadruped robots to successfully traverse challenging obstacles in various complex environments.

\end{abstract}


\section{INTRODUCTION}
The natural environment is extraordinarily complex, characterized by irregular terrains and unstructured obstacles in three-dimensional spaces. Humans and animals rely on their robust limbs to locomote through complex environments by running, climbing, jumping, and altering their body postures. This often necessitates real-time visual perception and depth of the surrounding environment to coordinate limbs to avoid obstacles. However, in the real world, both humans and animals have a limited field of vision, yet our limbs can move well beyond our visual perception range. When limbs move beyond the visual range or in environments where visual perception is ineffective (such as at night or in dense forests), perceiving environmental obstacles through proprioception becomes crucial. Designing a controller for robots that can navigate locomote through complex 3D environments and effectively avoid obstacles without relying on visual or lidar like external sensors presents a significant challenge.

In practical situations, humans and animals can rely on the sense of touch on their skin to detect the presence of obstacles around them. In robotic systems, this capability is referred to as collision detection. However, tactile sensors in robots are typically installed only in specific locations (such as the soles of the feet or fingertips), which results in the majority of the robot's body being unable to directly sense obstacles. Therefore, implementing collision detection for each limb becomes critically important. By relying on collision detection, robots can perceive unknown obstacles and guide their limbs movements, such as navigating through narrow spaces or finding paths in darkness. Ensuring that robots can effectively detect and respond to collisions not only enhances their autonomy but also ensures safety in interactions and operations within these unpredictable environments.


For quadruped robots, perceiving obstacles in complex three-dimensional environments poses a challenge. It requires three phases: detecting collisions, perceiving potential obstacles through the collisions, and replanning actions accordingly. Regarding collision detection, the current literature categorizes the entire collision process of robots into seven stages, referred to as the "collision event pipeline," including pre-collision, detection, isolation, identification, classification, reaction, and post-collision phases\cite{c1}. Among these, collision isolation specifically refers to locating the collision segment on the robot and its contact point. Most mature collision detection and isolation methods are designed for fixed-base robotic arms\cite{c2}, \cite{c3}, \cite{c4}, \cite{c5}, based on models or preset thresholds. Their primary aim is to halt the robotic arm in emergency situations to protect the operated object, the robotic arm itself, or human safety during human-robot interactions. However, these conventional methods are incapable of sensing potential obstacles generated from collisions, thus failing to redirect the robot's motion planning. Currently, end-to-end quadruped robot controllers based on reinforcement learning perceive complex terrains through proprioceptive observations, yet such methods are limited to terrain perception and do not extend to obstacle perception in three-dimensional environments\cite{c8,c9,RMA,Anymal2020sci,Anymal2019sci,shanghaiAI,TransferEXP, CrossLoco, FastMimic, QuadAMP, InstructionLearing}. Recently, some studies have used reinforcement learning to train quadruped robots for extreme parkour\cite{RobotParkourLearning}, \cite{ExtremeParkourLeggedRobots, Anymal2022sci, AnymalParkour, AnymalWheeled, AnymalDTC, quadABS, QuadEgo, VBC, AnymalLidar}, navigating through complex obstacles in three-dimensional spaces. However, these studies rely on external sensors, such as cameras, and have not proposed a solution that relies solely on proprioceptive sensing.

\begin{figure*}[h]
  \centering
  \includegraphics[width=\textwidth]{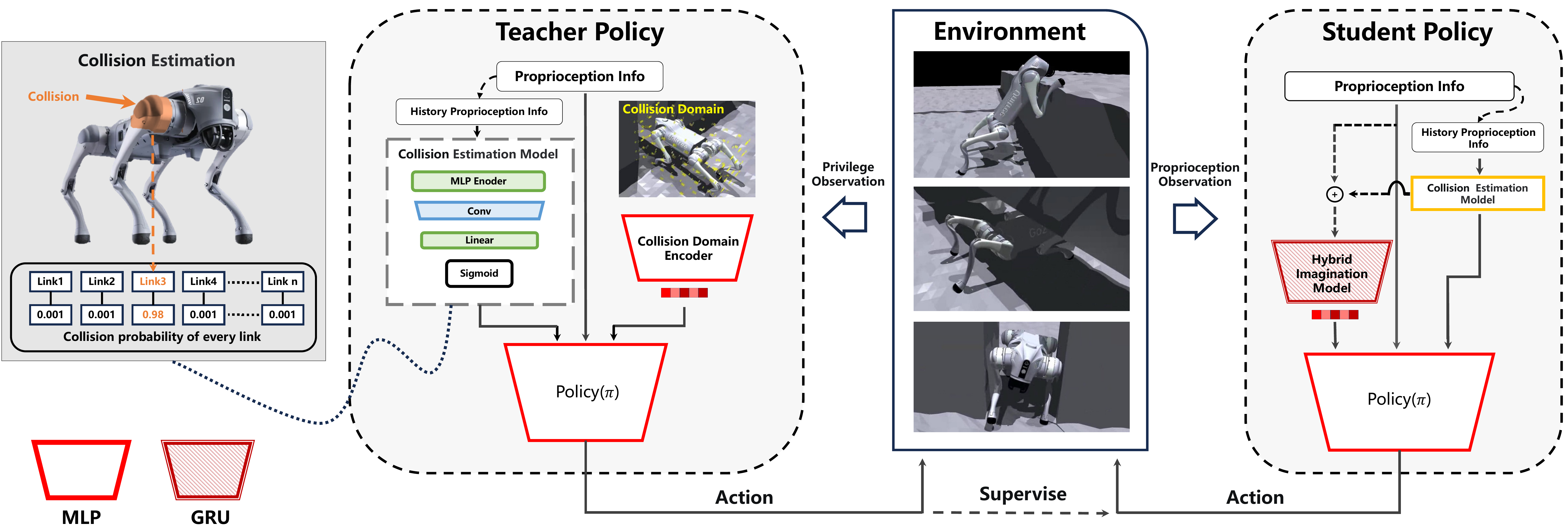}
  \caption{Teacher-student based two-stage training framework, where on the left is the schematic of the collision estimation.}
  \label{fig:wideimage}
  \vspace{-10pt}
\end{figure*}

\subsection{Related Works}

Existing collision detection methods can be primarily classified into model-based methods or model-free methods. \cite{c2}, \cite{c3}, \cite{c4}, \cite{c5} detect collision by comparing the estimated torques with predefined threshold. Among these, \cite{c2} proposes a momentum-based disturbance observer method. In comparison to the GM(generalized momenta-based) method, the MESO (modified extended state
observer) method introduced by \cite{c5} demonstrates improved collision force estimation under similar noise levels in practical systems. Both methods can locate the robot link with collision and provide directional information on the Cartesian collision force. However, in practical robot applications, these model-based methods require high precision in the robot's model and are sensitive to disturbances and loads.

During the collision isolation phase \cite{c1}, achieving more accurate localization of collision points/contact points relies on external sensors, such as those based on acoustics \cite{c6}. In model-free methods, approaches based on learning have also been proposed, but they necessitate external sensors such as laser rangefinders, cameras, and the collection of datasets \cite{c7}. The work discussed above is based on robotic arms with fixed bases. However, on robots with floating bases, the processes of collision detection, isolation, and identification become significantly more complex. This increased complexity arises because of more DOF, and the movement of these robots relies on continuous collision between their feet and the ground.

Recently, methods for collision detection, isolation, and identification in humanoid robots have been introduced \cite{c11, c12, c13}. In \cite{c13}, relying solely on proprioceptive sensing, the authors combine and compare the GM approach with an Force/Torque-sensor-based approach. However, these methodologies have predominantly been tested in simulated environments with flat terrain. In real-world applications with robots, additional challenges such as communication delays, sensor synchronization, and noise must be taken into account. In the realm of body collision detection for quadruped robots, prior model-based efforts have proposed various state estimation techniques for the robot\cite{c14}, \cite{c15}, \cite{c16}, attempting to fuse different sensors for more accurate state estimation \cite{c17}. Under conditions where only onboard perception is available, Fink et al. propose a sensor-less model based on kinematics to estimate the location of a single contact point at the shin level in real robot\cite{c10}.

In the domain of model-free methods, collision/contact detection primarily involves processing sensor signals in the frequency domain \cite{c18} or constructing classifiers using machine learning approaches \cite{c19}. In \cite{c20}, a deep learning-based contact estimator is developed using only onboard perception. However, these methods are predominantly applied to estimate foot forces for quadruped robots and do not consider the possibility of collisions at non-foot locations.  \cite{c8, c9} have trained implicit estimators within their policies, treating external forces which applied to the robot as privileged observations. This enables the use of historical data to inform subsequent movements. To the best of the authors' knowledge, there has yet to be research that trains an explicit estimator for collision detection and isolation, evaluates its accuracy in observing collisions, and utilizes this estimator to guide the robot's movement across varying environments.
\subsection{Contributions}
To further enhance the locomotive capabilities of quadruped robots without external perception, we have developed an end-to-end trained adaptive motion controller for quadruped robots, extending their mobility into complex three-dimensional environments. Our controller integrates proprioceptive sensing and collision estimation to implicitly imagine the features of obstacles in 3D spaces, enabling precise collision detection, localization of collision points, and agile collision responses and movements.

Our contributions are as follows:

\begin{itemize}
  \item We introduced a collision estimator that utilizes historical proprioceptive data to precisely estimate the likelihood of collisions on each body part, guiding the robot to develop targeted response strategies through neural networks.
  \item We developed the concept of a Collision Domain and a Hybrid Imagination Model to capture and estimate the characteristics of three-dimensional obstacles, enhancing our method's ability to classify diverse obstacles in complex settings.
  \item We established a two-phase end-to-end training framework for quadruped robots, employing a simple linear velocity reward with directional constraints. This framework, relying solely on proprioceptive sensing, enables agile movement in complex 3D environments and demonstrates more robust locomotive performance in simulations and real-world scenarios compared to baseline methods.
\end{itemize}

\section{Method}

\subsection{Task Formulation}
We define the process of a quadruped robot traversing a 3D complex environment without the assistance of external sensors as four stages: 1) encountering and colliding with obstacles, 2) locating the specific position of the collision, 3) estimating the encountered obstacle, and 4) making a swift response to overcome the obstacle. When traversing in 3D complex environments, quadruped robots may unexpectedly collide with obstacles. Due to the uncertainty of the contact points, we first developed a collision detection estimator to determine whether a collision has occurred and its specific location. Our method of estimating encountered obstacles distinguishes us from other studies\cite{RMA,Anymal2020sci}: unlike the traditional method of encoding terrain with elevation maps, we introduced the concept of a 3D collision domain, aimed at detecting obstacles that are about to come into contact with the robot's body. By combining the results of the collision detection estimator and the robot's proprioceptive abilities, we can effectively predict obstacles within the collision domain during actual movement. Based on this, the robot can take appropriate actions through real-time estimation of obstacles, effectively traversing various types of barriers.

\subsection{Base Set}

We model the environment as a Markov Decision Process (MDP). An MDP is defined by a tuple$(S, A, P_{a}, R_{a})$, where \(S\) represents the set of all possible states, and \(A\) represents the action space, \(P(s_{t+1}|s_t,a_t)\) is the state transition function, indicating the probability of transitioning to state \(s_{t+1}\) after taking action \(a_t\) in state \(s_t\), and \(R(s_{t+1}|s_t,a_t)\) is the immediate reward. The agent selects an action \(a_t\) from the policy based on the current state \(s_t\). For the state \(s_t\) and action \(a_t\), the MDP calculates the next state \(s_{t+1}\) and reward \(r_t\), and then provides feedback to the agent.  The goal is to select a policy that maximizes the cumulative sum of discounted rewards, expressed as: 

\begin{equation} 
J(\pi) = \mathbb{E}_{\pi} \left[ \sum_{t=0}^{\infty} \gamma^t r_t \right]
\end{equation}

\textbf{Policy Network} 
For the training of the policy $\pi_{\phi}(a_t | s_t)$, we employ an actor-critic architecture. In this framework, the actor improves its decision-making policy by maximizing the expected return estimated by the critic. We utilize the Proximal Policy Optimization algorithm(PPO) to efficiently optimize the policy.

\textbf{Action Space:} The output of the policy is a 12-dimensional tensor.  This tensor, once multiplied by a specific action scale coefficient $a_{scale}$, is amalgamated with a predetermined array of initial standing joint angles $\theta_{\text{default}}$. This process culminates in the formation of the targeted joint angles configuration:

\begin{equation}
\theta_{\text{desired}} = \theta_{\text{default}} + a_t \cdot a_{scale}
\end{equation}
Finally, proportional derivative (PD) controller is used for converting joint desired angle to joint torque.

\textbf{State Space:} For teacher policy, the state space $s_t$ is composed of proprioceptive observation $o_t$, explicit observations $\hat c_t$, $\hat v_t$, where $\hat c_t$ is the estimate of collision information, $\hat v_t$ is the estimate of body linear velocity, and implicit observations $p_t$, $e_t$, refer to the latent variables obtained after the 3D collision domain and privileged information are processed through the encoders.  Privileged implicit observation $e_t$ includes encoded body mass, center of mass, friction and motor strength. The state space $s_t$ is shown in the following equation:
\begin{equation}
s_t = \begin{bmatrix} o_t & \hat c_t & \hat v_t & p_t & e_t \end{bmatrix}^\mathrm{T}
\end{equation}
The structure of student policy is kept consistent with the teacher policy, where $e_t$ and $p_t$ are estimated using proprioceptive observations. Specifically, in the first phase, we employ Regularized Online Adaptation (ROA)\cite{ROA} to train a privileged information estimation module. In the second phase, we obtain estimates of privileged implicit information $\hat e_t$. Furthermore, in the second phase, we also train a hybrid imagination module to obtain estimates of $\hat p_t$, as detailed in section II.D.

\textbf{Reward:}
Without the need for prior knowledge from external sensors, previous work's reward mechanisms \cite{leggedgym,RobotParkourLearning,ExtremeParkourLeggedRobots} often cannot be directly applied to the tasks in this study. Our method primarily relies on observing the collision domain to master strategies for overcoming various types of obstacles. To ensure that the robot can smoothly traverse in the direction that obstacles are placed, we have designed an linear velocity tracking reward mechanism with heading constraints:
\begin{equation}
\ r_{vel} = L_{d_v} |\frac{min( v \cdot cos (\theta_{yaw}^{cmd} - \theta_{yaw}), v^{cmd})} { v^{cmd}}|
\end{equation}
$L_{d_v}$ is a weight factor that represents the positive and negative aspects of velocity, as follows:

\begin{equation}
L_{d_v} = \begin{cases} 
 K_{positive} &  ,v \cdot \cos (\theta_{yaw}^{cmd} - \theta_{yaw}) > 0 \\
 K_{negative} &  ,v \cdot \cos (\theta_{yaw}^{cmd} - \theta_{yaw}) < 0 
\end{cases}
\end{equation}

The linear velocity reward with heading constraint ensures that the robot moves along the direction of the obstacle placement, thus preventing the robot from choosing to bypass the obstacle. This design of $L_{d_v}$ stems from observations made during the training process: when faced with difficult obstacles, robots tend to adopt a policy of quickly bouncing back to their original position, then approaching the obstacle at a set speed and bouncing back again, which easily leads to the policy falling into a local optimum. Therefore, we included a penalty for negative velocity in the speed tracking reward to prevent the robot from adopting a rebound policy.Where both $k_{positive}$ and ${k_{negative}}$ are constant quantities, positive reward coefficient and negative penalty coefficient respectively, the detailed values are in the appendix.
To enable the robot to sidestep through narrow spaces and maintain a natural walking posture in normal environments, we introduced the following reward mechanism:
\begin{equation}
\begin{aligned}
\ r_{Pos} &= W_1 \cdot r_{GuidePos} + W_2 \cdot r_{Natural
Pos} \\
&= W_1 \cdot (|q_{hip}^{right} + q_{hip}^{left}|^2 + |q_{hip}^{front} - q_{hip}^{behind}|^2)  \\
& +W_2 \cdot |q_{dof}^{default} - q_{dof}|^2
\end{aligned}
\end{equation}
The pos reward consists of two parts, the GuidePos reward does not directly limit the angle of the robot's hip joint but penalizes the sum of the angles of the left and right leg hips to be zero and the difference in angles between the front and back leg hips, ensuring that the robot's limbs can always walk perpendicular to the ground under any circumstances. This not only facilitates the generation of sidestepping actions but the  NaturalPos reward ensures that the robot maintains a natural walking posture in normal environments. $W_1,W_2$ corresponds to the weights of the two-part rewards, the detailed values are in the appendix.

Although our method relies on collisions to perceive environmental obstacles, we have found that adding a collision penalty during actual training can accelerate the convergence of the policy. This is because, although collisions are necessary for perceiving obstacles, they should be avoided as much as possible during the traversal of obstacles. Therefore we refer to the collision penalty as follows, $\mathcal{F}_{collision}$ is the set of forces applied to the center of mass of each link.
\begin{equation}
\ r_{collision} = \mathcal{F}_{collision}
\end{equation}
Furthermore, we enhance the quadruped robot's overall mobility through an auxiliary reward mechanism refer to \cite{leggedgym}.

    

\par
\begin{figure}[ht]
\centering
\includegraphics[width=.25\textwidth]{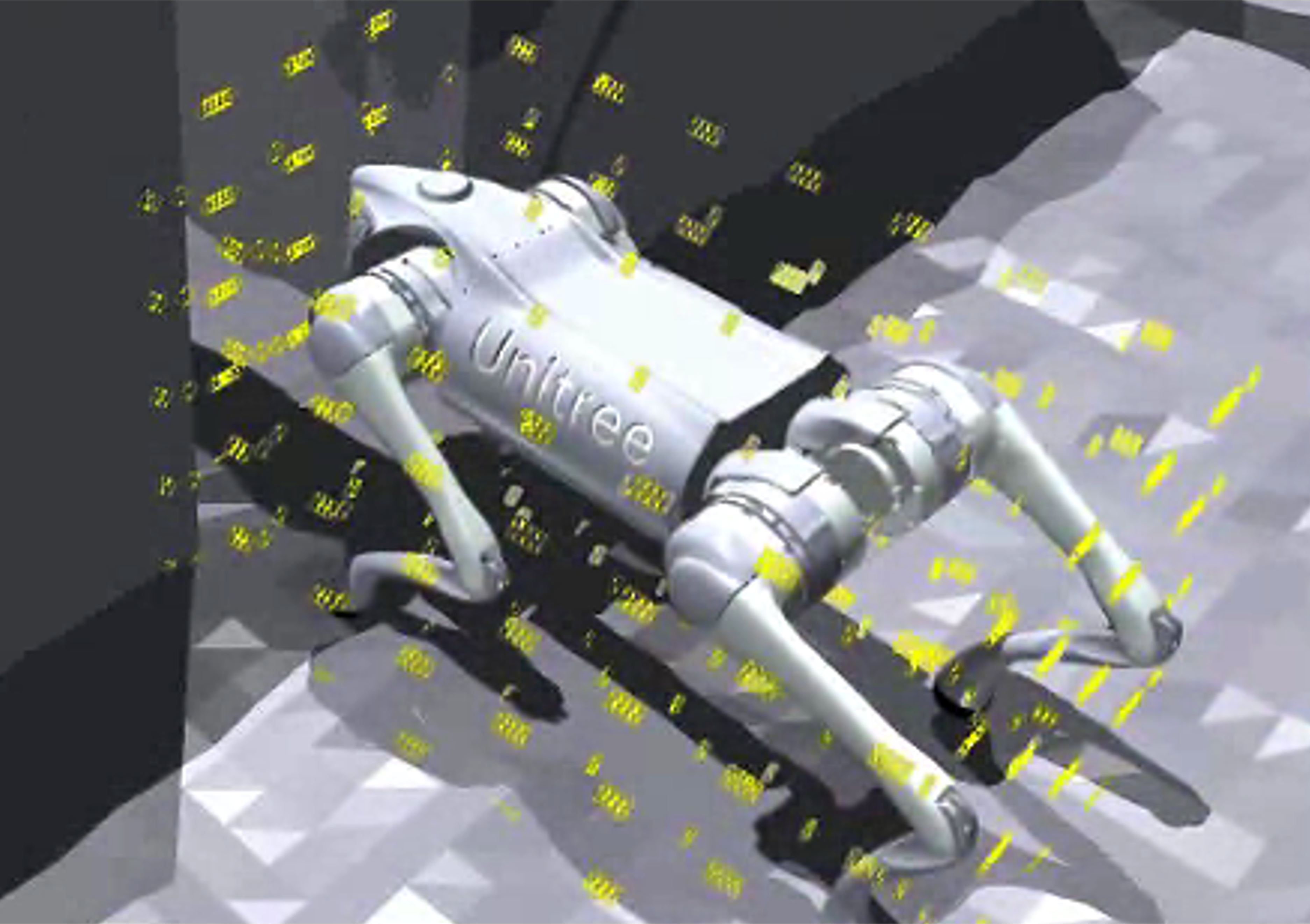}
\caption{Schematic of the collision domain}
\label{fig:wideimage}
\vspace{-10pt}
\end{figure}

\subsection{Collision Estimator}
In real-world applications, since robots cannot directly recognize collisions and their specific locations, we propose using a collision estimator $\varphi$ to estimate collision information online. In this study, collision information is simplified into a Boolean vector $c_t$, indicating whether each link has encountered a collision. In the simulation environment, we cannot directly obtain $c_t$ , so we determine that a collision has occurred if the force on each link exceeds a certain threshold. This threshold needs to be set according to different robots and the actual environment they are in. The aforementioned discussion indicates that most of the existing work on collision detection relies on dynamic information from a single timestep, which is effective for detecting collisions involving legs or arms. This is because collisions with legs and arms usually immediately reflect in the dynamics parameters, such as sudden changes in torque and acceleration, which can be captured through data within a single timestep. However, collisions involving the torso and joint parts are difficult to accurately capture due to the dispersed effects on dynamics parameters and response delays, and require analysis using long time series information. The collision estimator uses the robot's historical observation data to generate $\hat{c}_t$, which is the estimated value of the actual collision vector $c_t$.
\begin{equation}
\hat{c}_t = \varphi(o_{t-k}, o_{t-k+1}: o_t)
\end{equation}
Using sequence information can better estimate the source of collisions. In this paper, we used historical observation data from 10 timesteps. The length of the historical record should not be too long, as it may lead to overfitting. The selected length of the historical record should be sufficient to represent the sum of the occurrence of a complete collision event and the response time generated. The impact of history sequence on collision estimation is detailed in the experimental section.

Given that collision information is only divided into two states: collision occurred and collision did not occur, and the total probability of these two states sums to 1, we do not need to concern ourselves with a specific value vector, but only need to output a probability value. Therefore, the collision estimation problem can be simplified into a binary classification problem. Our method only requires the use of the current moment's collision state $c_t$ and its observation history, all of which can be obtained through simulation. The architecture of the collision estimator includes a linear layer, three convolutional neural network (CNN) layers, and another linear layer. The CNN layers apply convolution over the time dimension in the observation history to capture the temporal correlations in the input data. The flattened output processed by the CNN layers goes through another linear layer and applies a Sigmoid activation function, thereby producing an estimate of the collision state $\hat{c}_t$. This collision estimator is trained through supervised learning, with the goal of minimizing the following BCE loss function:
\begin{equation}
\mathcal{L}^{Est}_{\varphi}=-( c_tlog\hat{c}_t + (1 - c_t) log(1 - \hat{c}_t) )
\end{equation}

\begin{figure*}
    \centering
    \includegraphics[width=\textwidth]{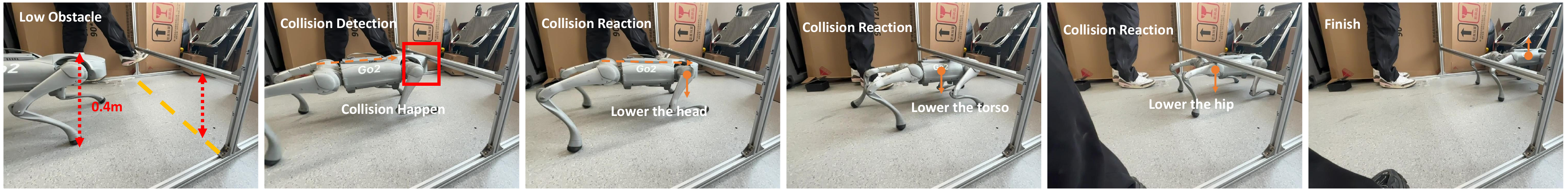}
    \caption{Locomotion through a tunnel: Adaptation strategies for collision avoidance.  }
  \label{fig:wideimage}
  \vspace{-10pt}
\end{figure*}

\vspace{-5pt}

\subsection{Implicit Collision Domain imagination}
To better perceive and respond to obstacles in the complex 3D environments, this study introduces the concept of a collision domain, which is an area around the robot's body containing information about obstacles, as shown in Fig. 3.  Specifically, we have set a cuboid sampling domain centered on the robot, see Table 1 for specific values. A boolean value indicates whether each point of the grid is inside an obstacle to detect imminent collisions.  In complex environments such as dense forests and low visibility areas where external sensors like lidar and cameras often fail, proprioception becomes the sole means of understanding the collision domain.  Therefore, this study proposes a method that uses only proprioception to deeply estimate the characteristics of the collision domain.

Previous research has shown that terrain characteristics can be estimated using only proprioception\cite{RMA,Anymal2020sci}.  In this study, we extend this idea to understanding features in 3D space, that is, the implicit imagination of the collision domain.  Here, we propose a two-stage training method that combines implicit collision domain imagination to develop a policy effective in traversing complex environments.  We combine proprioception with the aforementioned collision estimator to comprehensively estimate the potential attributes of the collision domain.  This approach is better suited to adapt to the cognitive uncertainties present in real-world scenarios than methods that directly input proprioceptive sensing.

The specific training framework of our method is shown in Fig. 2. In the first phase, we use an encoder to extract latent features from the collision domain, inputting the resulting latent variable $p_t$ directly into the teacher policy to train the robot. In the second phase, we supervise the training of the student policy using the teacher policy. Since the student policy cannot directly access $p_t$, we propose a hybrid imagination model that combines proprioceptive observations $o_t$ with estimates $\hat c_t$ from the collision estimator, feeding them into a GRU pipeline to estimate latent features of the collision domain. The advantage of the hybrid imagination model is that, after embedding explicit collision estimates, the GRU module's temporal feature extraction mechanism can accurately represent the process of the robot interacting with obstacles. This allows for a dynamic estimation of environmental obstacle information, further estimating the latent features of the collision domain, thereby improving the robot's response capability after encountering obstacles.

\section{EXPERIMENTS}


\begin{table}
    \setlength\tabcolsep{6pt}
    \fontsize{10}{10}\selectfont
    \centering
    \caption{Curriculum Learning: Obstacle Configuration and Robot Parameter Settings.}
    \label{tab:my_label}
    \begin{tabular}{ccc} 
        \toprule
        \multicolumn{3}{c}{Obstacle  }\\ 
        
        Properties& Train Ranges (m) &Test Ranges (m) \\
        & ([$l_{easy}$, $l_{hard}$])&([$l_{easy}$, $l_{hard}$])\\ 
        \midrule\\

         Highland&  [0.05, 0.55]&  [0.25, 0.55] 
\\ 
         Barrier&  [0.31, 0.00]&  [0.16, 0.00] 
\\ 
         
Tunnel&  [0.40, 0.25]&  [0.38, 0.25] 
\\ 
         Crack&  [0.38, 0.28]&  [0.32, 0.28] \\ 
 \bottomrule
 \toprule
 \multicolumn{3}{c}{Parameters}\\ 
 Properties& Go2 Body (m) &Collision Domain (m) 
\\ 
\midrule\\
 length& 0.71&0.90 
\\ 
 width& 0.32&0.40 
\\ 
 height& 0.40&0.50 \\ 
\bottomrule
    \end{tabular}
  \vspace{-10pt}  
\end{table}

\subsection{Training set up} 
The simulator utilized for training the policy is Isaac Gym\cite{isaacgym}, supplemented by the prior open-source libraries Legged Gym and RSL-RL\cite{leggedgym} for. We conduct parallel training across 4,096 domain-randomized environments on a single NVIDIA RTX 4090 GPU, teacher policy engaging in approximately 12000 training iterations and student policy engaging in not more than 10000 training iterations, each comprising 24 steps, with the policy operating at a control frequency of 50Hz.

To accurately reflect the diversity of obstacles encountered in the real world, we categorize four types of obstacles, which included: Highlands, which are climbable terrains directly ahead, as shown in Fig. 5.; Barriers in Fig. 7, obstacles located on the left or right side of the robot; Tunnels, low tunnel-like obstacles that the robot can pass through by lowering its body height like Fig. 4; and Cracks, narrow passages shown in Fig. 7. Additionally, we implement a strategic obstacles curriculum designed to enhance the policy's generalization capabilities and convergence rate.

\begin{table*}

    \centering
    \caption{We test our method against several baselines and ablations in the simulation with obstacle of varying difficulty (see in Table I). We initialized a total of 4096 robots and randomly distributed them across the map . We collected data on the success rate (the ratio of the number of robots that failed to traverse obstacles to the total number of robots) and the average movement distance which was normalized to [0, 1] after a duration of 10 seconds.}
    \label{tab:my_label}
    \setlength\tabcolsep{11pt}
    \fontsize{10}{12}\selectfont
    \begin{tabular}{c|cccc|lllc}
        \toprule
         &  \multicolumn{4}{c|}{Success Rate  ↑}&  \multicolumn{4}{c}{Average Displacement↑}\\
 & Highland&Barrier&Tunnel&Crack& Highland& Barrier& Tunnel&Crack\\
 \midrule
         Baseline&  0.00&   0.12&  0.00&  0.00&  0.004& 0.132& 0.005&0.006\\
         RMA&  0.27&  0.61&  0.67&  0.53&  0.256& 0.550& 0.614&0.497\\
         WTW&  0.00&  0.11&  1.00&  0.00&  0.004& 0.103& 1.000&0.005\\
 Ours w/o R.V& 0.00& 0.31& 0.30& 0.00& 0.003& 0.197& 0.218&0.005\\
 Ours w/o Col& 0.89& 0.81& 0.89& 0.76& 0.769& 0.771& 0.898&0.625\\
 Ours w/o H.O& 0.93& 0.87& 0.94& 0.83& 0.796& 0.793& 0.959&0.711\\
 Ours& 0.94& 0.92& 0.96& 0.89& 0.821& 0.853& 0.979&0.798\\
 \midrule

 Teacher& 0.99& 1.00& 1.00& 1.00& 1.000& 1.000& 1.000&1.000\\
 \bottomrule
    \end{tabular}

\end{table*}

\subsection{Hardware and Depoly} 
In our research, we employ the Unitree Go2 robot to assess the efficacy of the collision detection estimator and the policy's response to collisions. The body parameters of Go2 are shown in Table 1, which serves as an important reference for setting up our training environment. For the deployment of policies on real robots, these policies are executed on the onboard NVIDIA Jetson Orin Nano of the Go2. Furthermore, \cite{walk-these-ways} provides an exemplary framework that facilitates the straightforward transplantation of these policies to the Go2 robot.

\begin{figure}[ht]
  \vspace{10pt}
    \centering
    \includegraphics[width=.5\textwidth]{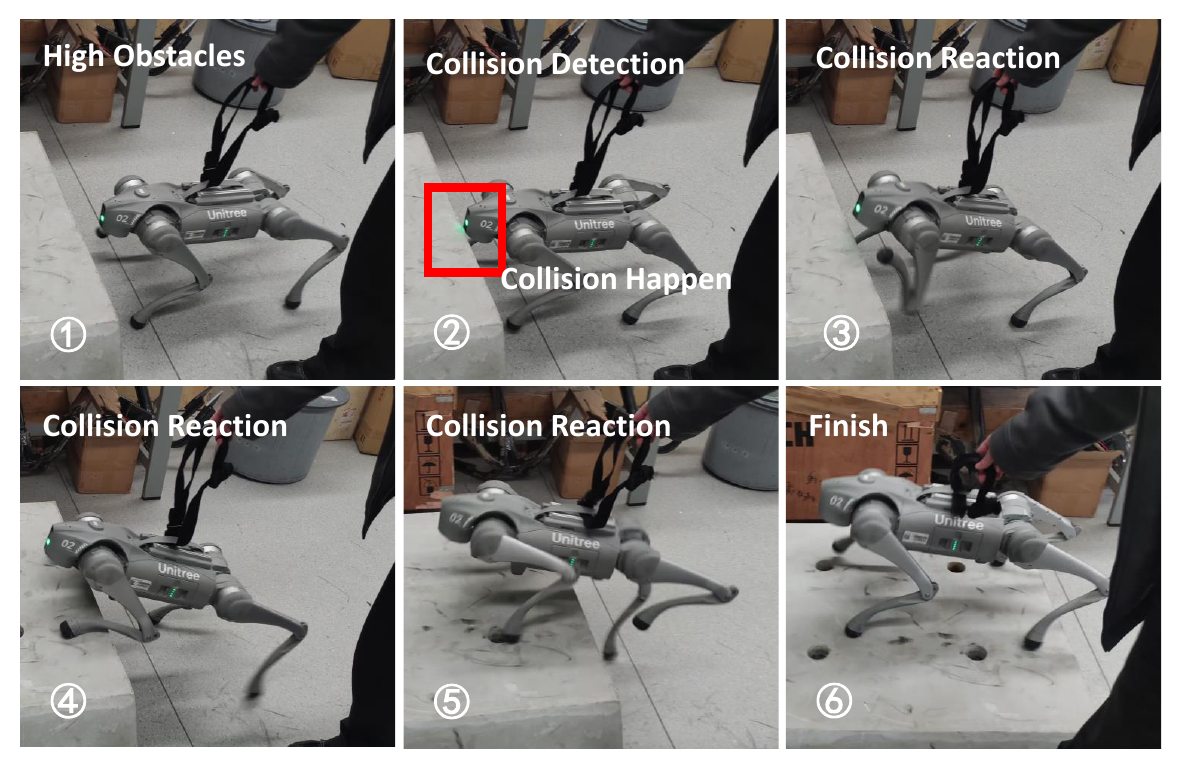}
    \caption{Collision response strategies when facing a highland.}
  \label{fig:wideimage}
  \vspace{10pt}
  \end{figure}
\subsection{Compared Method}

We compare our collision estimation and response policy with several baseline and ablations as follows.

\vspace{2mm}

\begin{itemize}
    \item \textbf{Ours w/o R.V:} We use the ordinary linear velocity tracking reward \cite{leggedgym} instead of the linear velocity tracking reward with heading constraints.
    \item \textbf{Ours w/o Col:} Training without collision estimator.
    \item \textbf{Ours w/o H.O:} Training collision estimator without history observation.
    \item \textbf{Baseline:} Training directly with only proprioception.
    \item \textbf{RMA\cite{RMA}:} Employing an Adaptation Module to Estimate All Privileged Observations Including Terrain Height from Historical Data, Harnessing Potential Information.
    \item \textbf{WTW(Walk-These-Ways)\cite{walk-these-ways}:} Leverages expert knowledge realize Multiplicity of Behavior.
    \item \textbf{Go2-default:} Go2 default controller based on NMPC, which only compared in real experiment.
    \item \textbf{Go2-special:} Go2 special controller based on NMPC, which only compared in real experiment.

\end{itemize}

\subsection{Simulation Experiments} 
\textbf{Baselines and Ablations.}
We conducted a comparative analysis of our method against baseline approaches such as RMA and WTW, which are two typical methods in this domain. The RMA approach extracts information from 2D elevation maps and privileged data to adapt to the environment, whereas WTW relies on expert knowledge to set rewards and manually adjust the robot's posture for obstacle traversal. In addition to RMA's framework, we implemented our reward function. As shown in TABLE II, the 2D elevation maps used in RMA struggle to capture the complete features of 3D obstacles in the environment, resulting in significantly lower performance across multiple tasks compared to our method. WTW can control the robot's height, which allows the robot to crawl to traverse Tunnel obstacles but lacks the ability to climb or sidestep, failing to navigate complex obstacles like Highland, Crack, and Barrier.

Comparing our method with Ours w/o R.V, we found that the velocity reward without heading direction constraint failed to traverse Highland and Crack obstacles, and also performed poorly with Barrier and Tunnel. This is because the previous method's velocity reward was solely dependent on the fixed-base linear velocity, leading robots to quickly circumvent rather than traverse obstacles, a common policy during training due to the lower exploration difficulty of avoiding obstacles.

The comparison between Ours and Ours w/o H.O highlights the importance of historical information for collision estimation, particularly when traversing Barrier and Crack obstacles. These obstacles often involve more collisions at the hip joint, and estimating collisions from a single time-point is challenging. Furthermore, the comparison suggests that the inclusion of collision information significantly improves the robot's performance in navigating various complex obstacles. Collision information aids the robot in inferring the characteristics of the obstacle from the contact, enabling further appropriate responses.

\begin{figure}[ht]
\vspace{10pt}
\centering
\includegraphics[width=.5\textwidth]{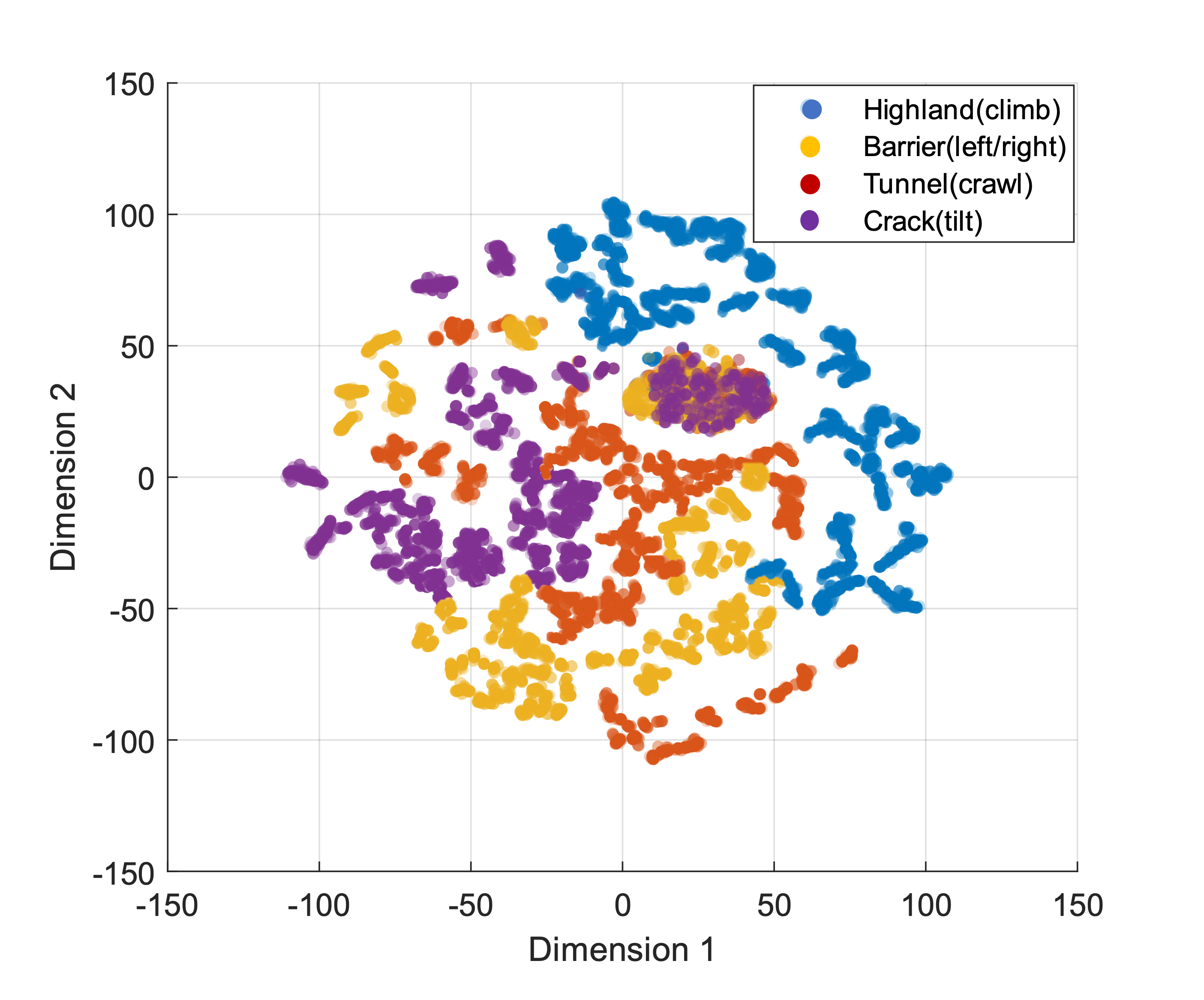}
\caption{The t-SNE visualization for Collision Domain in the latent space.}
\label{fig:wideimage}
\vspace{10pt}
\end{figure}

\textbf{Effects of Implicit Collision Domain imagination. }
To thoroughly investigate the impact of Implicit Collision Domain imagination on the task and to quantify the efficacy of the Hybrid Imagination Model, we employed the t-distributed stochastic neighbor embedding (t-SNE) to reduce the dimensions of the output from the Hybrid Imagination Model. As illustrated in the Fig. 6., we discovered that the latent representation of the Collision Domain estimated by the Hybrid Imagination Model shows a clear distribution across different obstacle categories. This indicates two points: 1. The Collision Domain can effectively extract features of obstacles in complex three-dimensional environments. 2. The Hybrid Imagination Model can efficiently estimate the latent representation of the Collision Domain, thereby possessing sufficient environmental information to aid the robot in responsive decision-making in complex settings. Additionally, the t-SNE plot shows that some overlap between the Barrier and Crack obstacles, suggesting that the body locations where the robot collides with these two types of obstacles are often similar, which also indirectly shows that the latent representation of the Collision Domain accurately reflects the actual situation.



\subsection{Real-World Experiments}

\begin{figure}[!h]
    
    \centering
    \includegraphics[width=.5\textwidth]{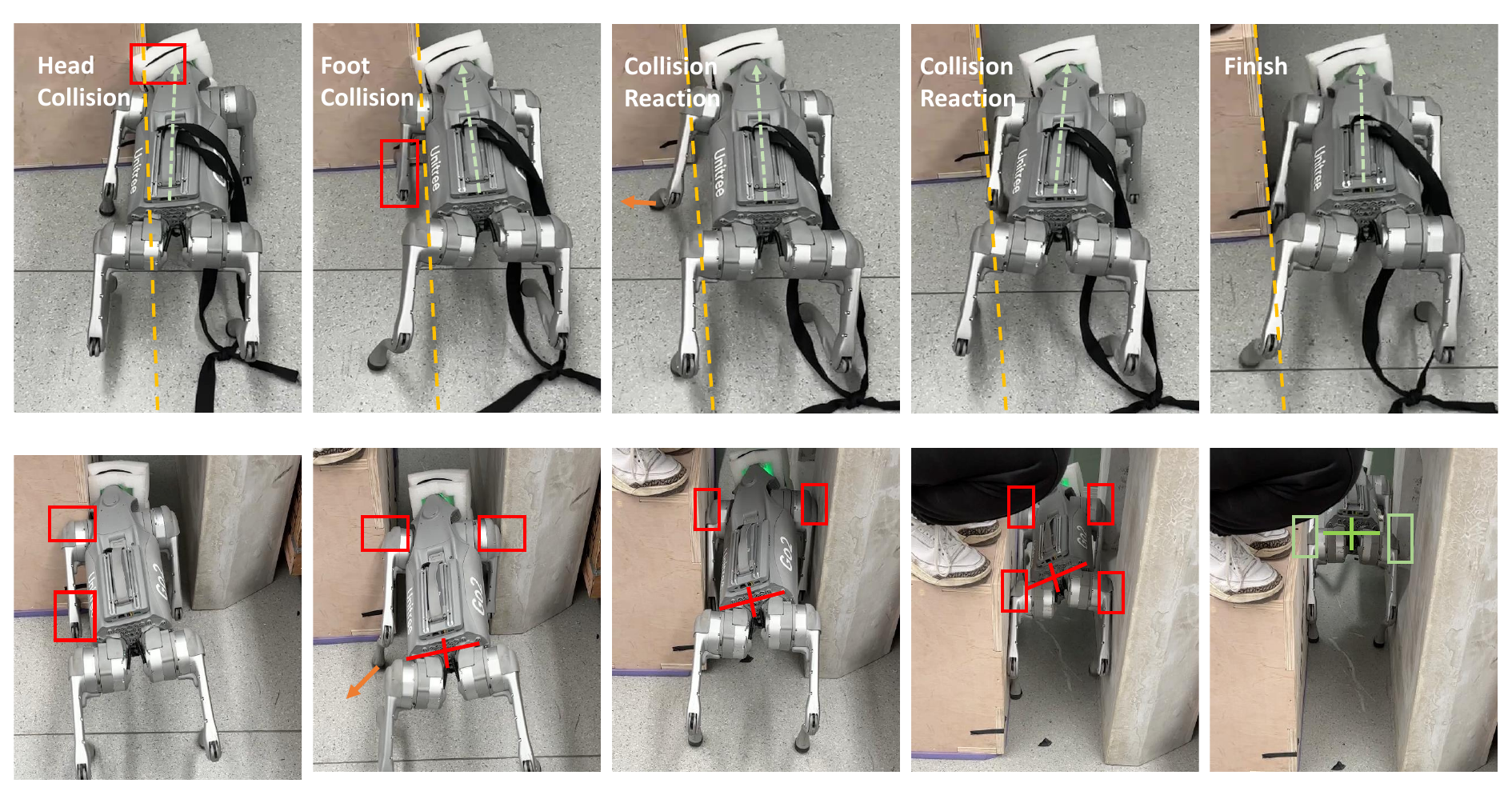}
    \caption{Collision response strategies when facing a barrier (first row) and crack obstacles (second row).}
    \label{fig:wideimage}
\end{figure}

\begin{figure*}[pt]
  \centering
  \includegraphics[width=\textwidth]{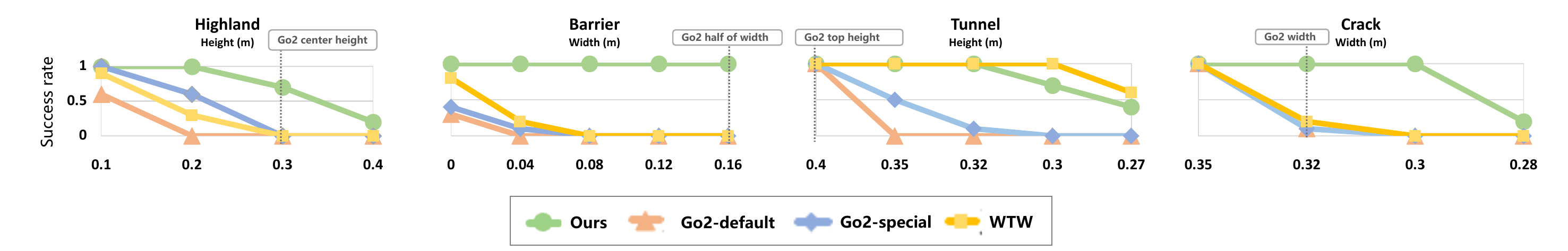}
  \caption{Real-world quantitative experiments.}
  \label{fig:wideimage}
\end{figure*}

\textbf{Qualitative Experiments.}
In an indoor environment, we conducted a series of experiments aimed at exploring different obstacles through collision detection including highlands, barriers, tunnels, cracks to qualitatively analyze the impact of collisions on a robot's ability to traverse unknown obstacles.In exploring the Highlands,as shown in Fig. 5., the robot first used its head collision detection to perceive potential highland obstacles. Subsequently, the robot extended its right front foot for further exploration. Upon detecting a collision with the right front foot, the robot determined the presence of a highland obstacle and chose to overcome it by climbing. When facing a Barrier, the robot detected the obstacle through a collision on the left side of its head. Then, by probing with its front feet, it confirmed the obstacle was on the left while the right side was clear, thus, the robot chose to turn right to bypass the obstacle, as shown in Fig. 7.. In the Tunnel obstacles, after an initial head collision detection, the robot did not detect any horizontal collisions with its hands, leading to the determination that the obstacles was a tunnel. The robot then successfully passed through by lowering the height of its torso and hips, as shown in Fig. 4. In the exploration of Crack obstacles, the robot's left front hip and left front foot first detected a collision. Attempting to move right, the robot encountered another collision at the right front hip, thereby determining the obstacles to be a crack. The robot chose to adjust the roll axis angle of its body to navigate through the obstacle, as shown in Fig. 7.

Additionally, we tested our approach in complex wild scenarios, as depicted in Fig. 1., where the robot traverse unstructured rocky obstacles.  This presents a significant challenge as both collision points and potential footholds are unpredictable, making it difficult for the robot to identify obstacle characteristics.     To our knowledge, there have been almost no other works that have achieved similar results. But our team managed to traverse wild scenarios which has various obstacles without external sensors, achieving a success rate of over 50\%. More details on these wild experiments are available in the accompanying video.

\textbf{Quantitative Experiment.}
As depicted in the Fig. 8., we have quantitatively compared our method with three other strategies: Go2 Default, Go2 Special, and WTW, in real-world scenarios. The data shown represents the average success rate of the robot when confronting various types of obstacles at different levels of difficulty, across 10 trials, aiming to showcase the obstacle negotiation capability of our approach in actual environments. It is clear from the figure that experiments utilizing our policy exhibited superior performance in almost all obstacle types. Notably, in the specific scenario of navigating through Tunnels, the WTW policy performed slightly better, primarily due to its use of human knowledge priors and manual adjustment of the robot's height for successful passage. In contrast, our method is adaptive, which may lead to certain errors in estimating environmental obstacles, resulting in a degree of performance degradation. However, this performance degradation is within an acceptable range.


\section{CONCLUSION}

We present a novel quadruped robot collision response motion controller based on collision domains and explicit collision estimator, capable of precise collision detection, localization, and agile response in unknown and complex 3D environments, solely relying on proprioception. Comprehensive evaluations in both simulated and real-world environments have demonstrated the precision of our collision detection technique, the robustness of our collision response, and the effective traversability in complex environments.However, a limitation of this work is that during the simulation training process, our approach necessitates a relatively accurate robot collision model, as it significantly dictates the robot's capability to perceive environmental obstacles. However, crafting such an accurate collision model in the simulation phase demands substantial computational resources, and the calculation of collision contact forces in complex models introduces certain errors. We look forward to finding a solution to this challenge in the future. Additionally, in our forthcoming research, we aim to integrate a recognition network to enhance the accuracy of estimating both the direction and magnitude of forces. This will enable us to precisely mitigate potential disturbances and implement adaptive control across environments characterized by diverse stiffness and contact elasticity.

\clearpage

\def\maketitlesupplementary
   {
   \newpage
       \oldtwocolumn[
        \centering
        \Large
        \textbf{Appendix}\\
        \normalsize
        \vspace{0.5em} 
        \vspace{1.0em}
       ] 
   }
\maketitlesupplementary

\subsection{Reward Function}
Reward items are listed in TABLE III. The task reward helps the robot adapt to the specific task scenarios, and the details of the design (4,6,7) is shown in II.B above. Where we set the Goal Velocity's reward coefficient $K_{positive}$ to 1 and the penalty coefficient $K_{negative}$ to -3. This set of parameters was optimized based on our comprehensive simulation experiments. Additionally, we designed the regularization reward referring to \cite{leggedrobot}, adjusted it according to the type of robots and task scenarios.

\begin{table}[h!] 
    \centering
    \caption{Reward Function}
    \label{tab:my_label}
    \setlength\tabcolsep{9pt}
    \fontsize{6}{10}\selectfont

    \begin{tabular}{ccc}
        \toprule
         Term&  Equation& Weight\\
         \midrule
         \multicolumn{3}{c}{Task reward} \\ 
        \midrule
         Goal Velocity&  $r_{vel}(4)$ & 1.5\\
        Yaw Angular Velocity&  $exp(-4|\omega^{cmd}_{yaw}-\omega{yaw}|)$& 0.5\\
         Hip Position& $r_{Pos}(6) $ &-0.5($W_1$=$W_2$=0.5)\\
         Collision&  $r_{collision} (7)$& -10.0\\
        \midrule
         \multicolumn{3}{c}{Regularization reward} \\ 
        \midrule
        Z Velocity&  $v_z^2$& -0.5\\
         X\&Y Velocity&  $\omega_x^2+\omega_y^2$& -0.01\\
         Dof Acceleration&  $\sum_{i=1}^{12} \ddot{q}_i^2$& $-2.5*10^{-7}$\\
         Action Rate& $\sqrt{\sum_{i=1}^{12} (a_t - a_{t-1})^2}$&-0.1\\
         Delta Torques& $\sum_{i=1}^{12} (\tau_t - \tau_{t-1})^2$&$-1.0*10^{-7}$\\
         Torques& $\sum_{i=1}^{12} (\tau_t)^2$&$-1.0*10^{-5}$\\
         Dof Error& $\sum_{i=1}^{12} (q - q_{default})^2$&-0.04\\
         Feet Stumble& $|F_{feet}^{hor}|>4*|F_{feet}^{ver|}$&-1\\
         Dof Position Limits& $\sum_{i=1}^{12} ({q^{out}_i}_{q_i > q_{\text{max}} \|\, q_i < q_{\text{min}}})$&-10.0\\
         \bottomrule
    \end{tabular}

\end{table}

\subsection{Additional Training Details}
\justify

\noindent \textbf{Network Architecture}
Our learning framework's overall network architecture includes not only the components shown in Figure 2 but also some additional modules not depicted in the figure. The teacher policy consists of six Multilayer Perceptron (MLP) parts: collision domain encoder, collision estimator, privileged information encoder, privileged information estimator, velocity estimator, and teacher policy network. Where $\boldsymbol{h}_t$ denotes collision domain information and $\boldsymbol{g}_{t}$ denotes privilege information. The privileged information encoder supervises the privileged information estimator and optimizes learning using the ROA method. Privileged Estimator's network type is also a CNN with a similar structure to the Collision Estimator. The student policy includes five parts: the collision estimator, velocity estimator, and privileged information estimator from the teacher policy, along with the hybrid imagination model comprised of a Gated Recurrent Unit (GRU) network and an MLP-based student policy network. Table I provides more details on each layer.

\begin{table}[h]
\setlength{\abovecaptionskip}{0.cm}
\setlength{\belowcaptionskip}{-0.cm}
\centering
\caption{Network architectures}
\label{tab:net_arc}
\setlength\tabcolsep{6pt}
\fontsize{6}{10}\selectfont
\begin{tabular}{@{}lllll@{}}
\toprule
\textbf{Module} & \textbf{Inputs} & \textbf{Hidden Layers} & \textbf{Outputs} \\ \midrule
\multicolumn{3}{c}{Teacher policy} \\ 
\midrule
Collision Domain Enc &$\boldsymbol{h}_t$&{[}256, 128, 64{]}& $\boldsymbol{p}_t$ \\
Collision Estimator& $\boldsymbol{o}_{t-10} ,...,\boldsymbol{o}_{t-1} ,\boldsymbol{o}_{t} $    & /            & $\hat{\boldsymbol{c}}_t$  \\
Privileged Encoder& $\boldsymbol{g}_{t}$   & {[}64, 32{]}  & $\boldsymbol{e}_t$ \\
Privileged Estimator& $\boldsymbol{o}_{t-5} ,...,\boldsymbol{o}_{t-1} ,\boldsymbol{o}_{t} $ & /  & $\hat{\boldsymbol{e}}_t$ \\
Velocity Estimator&$\boldsymbol{o}_{t} $ & {[}128, 64{]}     & $\hat{\boldsymbol{v}}_t$  \\
Teacher Network&$\boldsymbol{o}_t,\boldsymbol{p}_t,\hat{\boldsymbol{c}}_t,\boldsymbol{e}_t,\hat{\boldsymbol{v}}_t$ & {[}512, 256, 128{]}     & $\boldsymbol{a}_t$  \\
\midrule
\multicolumn{3}{c}{Student policy} \\ 
\midrule
Hybrid Imagination Model&$\hat{\boldsymbol{c}}_{t} ,\boldsymbol{o}_{t}$&[128, 64]& $\hat{\boldsymbol{p}}_{t}$  \\
Student Network& $\boldsymbol{o}_t,\hat{\boldsymbol{p}}_{t},\hat{\boldsymbol{c}}_{t},\hat{\boldsymbol{e}}_{t},\hat{\boldsymbol{v}}_{t}$ & {[}512, 256, 128{]}& $\hat{\boldsymbol{{a}}}_t$ \\ \bottomrule
\end{tabular}
\end{table}

\noindent \textbf{Training course}
Course learning is crucial for robots to effectively travel obstacles in complex environments.  Without this capability, robots would struggle to learn effectively.  We have implemented the velocity-travel course training method\cite{AMP,leggedrobot}, where the robot's linear velocity is randomly sampled within the range of [0, 1].  If the robot's travel distance in one iteration exceeds half of the preset heading speed integral, the terrain difficulty is increased;  otherwise, it is decreased.  The levels of terrain difficulty are detailed in TABLE I.

\begin{table}[h]
        \centering
        \raisebox{-0.0\height}{
        \small
        \begin{tabular}{@{}lllll@{}}
        \toprule
        Hyperparameter & Value \\ [0.5ex]
        \midrule
         Discount Factor & $0.99$ \\ [0.5ex]
         GAE Parameter & $0.95$ \\ [0.5ex]
          Timesteps per Rollout & $21$ \\ [0.5ex]
          Epochs per Rollout & $5$ \\ [0.5ex]
          Minibatches per Epoch & $4$ \\ [0.5ex]
          Entropy Bonus ($\alpha_2$) & $0.01$ \\ [0.5ex]
          Value Loss Coefficient ($\alpha_1$) & $1.0$ \\ [0.5ex]
          Clip Range & $0.2$ \\ [0.5ex]
          Reward Normalization & yes \\ [0.5ex]
          Learning Rate & $2\mathrm{e}{-4}$ \\ [0.5ex]
          \# Environments & $4096$ \\ [0.5ex]
          Optimizer & Adam \\ [0.5ex]
         \bottomrule
        \end{tabular}
        }
        \vspace{0.45cm}
        \captionof{table}
          {%
            PPO hyperparameters. 
        \label{tbl:ppo_hparams}
          }
\end{table}

\noindent \textbf{Policy training and Imitation training}
We use PPO with hyperparameters listed in TABLE V to train the teacher policy. We regard the process of teacher policy supervising student policy learning as imitation learning process, \( a_{\text{t}} \) and \( \hat{a}_{\text{t}} \) are the action vectors from the teacher and student respectively in an actor network. The gradient of the loss \( \mathcal{L} \) can be defined as the sum of the squared norms of the difference between \( a_{\text{t}} \) and \( \hat{a}_{\text{t}} \) over all actions:
\[
\mathcal{L}_{Imitation} =  \left\| \pi_{\text{teacher}}(\cdot |s) - \pi_{\text{student}}(\cdot |s) \right\|_2^2
\]
Here, \( \left\| \cdot \right\|_2 \) denotes the \( L^2 \) norm. Where $\pi_{\text{teacher}}(\cdot |s)$ is the action from the Teacher policy and $\pi_{\text{student}}(\cdot |s)$ is the action from the Student policy .
\end{document}